\documentclass{article}

 \usepackage[preprint]{neurips_2020}

\usepackage[T1]{fontenc}    
\usepackage{hyperref}       
\usepackage{url}            
\usepackage{booktabs}       
\usepackage{amsfonts}       
\usepackage{nicefrac}       
\usepackage{microtype}      
\usepackage{array}
\usepackage{multirow}
\usepackage{amsmath}
\usepackage{amssymb}
\usepackage{graphicx}
\usepackage{subfig}

\title{Conditional Hybrid GAN for Sequence Generation}
\author{%
  Yi Yu \\
  National Institute of Informatics, Tokyo, Japan\\
  \texttt{yiyu@nii.ac.jp} \\
   \And
   Abhishek Srivastava\thanks{Abhishek Srivastava was involved in this work from April to May 2020 during his internship at the National Institute of Informatics, Tokyo, Japan.} \\
  MIDAS, IIIT-Delhi, India\\
  \texttt{abhishek18124@iiitd.ac.in} \\
   \AND
   Rajiv Ratn Shah \\
   MIDAS, IIIT-Delhi, India \\
   \texttt{rajivratn@iiitd.ac.in} \\
}

\begin{document}

\maketitle

\begin{abstract}
Conditional sequence generation aims to instruct the generation procedure by conditioning the model with additional context information, which is a self-supervised learning issue (a form of unsupervised learning with supervision information from data itself). Unfortunately, the current state-of-the-art generative models have limitations in sequence generation with multiple attributes. In this paper, we propose a novel conditional hybrid GAN (C-Hybrid-GAN) to solve this issue. Discrete sequence with triplet attributes are separately generated when conditioned on the same context. Most importantly, relational reasoning technique is exploited to model not only the dependency inside each sequence of the attribute during the training of the generator but also the consistency among the sequences of attributes during the training of the discriminator. To avoid the non-differentiability problem in GANs encountered during discrete data generation, we exploit the Gumbel-Softmax technique to approximate the distribution of discrete-valued sequences. Through evaluating the task of generating melody (associated with note, duration, and rest) from lyrics, we demonstrate that the proposed C-Hybrid-GAN outperforms the existing methods in context-conditioned discrete-valued sequence generation.
\end{abstract}

\section{Background and motivation}

Conditional sequence generation has been a challenging research task in the ﬁeld of artiﬁcial intelligence, which falls under the field of conditional discrete sequence generation. This generation aims to develop generative models that can automatically predict sequence with given context information in a way similar to the creativity of human. An earlier study [1] has shown the feasibility of exploiting conditional LSTM-GAN for sequence generation with multiple attributes. Although this state-of-the-art generation has demonstrated meaningful results compared with the traditional maximum likelihood estimation (MLE), it fails to accurately model the discrete attributes. On the one hand, the continuous-valued sequence as the output of the generator in the GAN, is not in accordance with the discrete-valued attributes. On the other hand, due to the quantization error, the generated attributes could be associated with an improper discrete-valued music attribute, which would lead to a negative impact on sequence generation.

To overcome the aforementioned disadvantage, in this work, the Gumbel-Softmax is exploited to approximate the distribution of discrete-valued sequences. On this basis, a novel conditional hybrid generative adversarial network(C-Hybrid-GAN) is suggested to generate discrete-valued sequences, where three discrete sequences of attributes are separately generated based on conditioning the same context. We evaluate our generation model utilizing paired melody-lyrics sequences in [1]. In particular, the relational reasoning technique is exploited to model the dependency inside each sequence of music attribute during the training of the generator as well as the consistency among three sequences of music attributes during the training of the discriminator. Through the extensive experiments, we conclude the proposed C-Hybrid-GAN outperforms the existing melody generation methods from lyrics through the extensive experiments and has the capability of generating more natural and plausible melodies.

\vspace{-0.4em}

\section{Related works}
Conditional sequence generation remains a challenging task, aiming to imitate the real sequence conditioned on the specific context input. In this work, we focus on exploiting GAN but having two significant contributions: i) relational reasoning technique is applied to modeling both dependency inside each sequence of the attribute and consistency among three sequences of the attributes during the training stage. ii) Gumbel-Softmax technique is utilized to approximate the discrete-valued distribution of attributes.
Generative adversarial networks (GANs) were originally developed to generate continuous data, which have been applied successfully in the conditional sequence generation such as lyrics-to-melody [1], text-to-video [2], and dialogue [3] generation. However, GANs have the limitation in generating discrete sequence due to the non-differentiable problem of the discrete-valued outputs from the generator. To overcome this disadvantage, existing works pay attention to two research lines: i) policy gradient based on reinforcement learning and ii) continuous approximation of the discrete distribution.

i) policy gradient based on reinforcement learning.
In MaskGAN [4] the authors propose the actor-critic GAN architecture that uses reinforcement learning to train the generator, where the in-filling technique may alleviate mode collapse. RankGAN [5] uses the ranking score as the rewards to learn the generator, which is optimized through the policy gradient method. LeakGAN [6] addresses a mechanism of providing richer information from the discriminator to the generator by exploiting hierarchical reinforcement learning. SeqGAN [7] models the generator as a stochastic policy in reinforcement learning, which avoids the generator differentiation problem by directly performing policy-gradient updates.

ii) continuous approximation of the discrete distribution.
Instead of applying standard GAN objective, FM-GAN [8] suggests to match the latent feature distributions of real and synthetic sentences exploiting the feature-movers distance. In TextGAN [9], the authors utilize a kernelized discrepancy metric to map high-dimensional latent feature distributions of real and synthetic sentences, with the aim of mitigating the model collapse. In ARAE [10], the authors utilize the adversarial autoencoder to transform the discrete data into a continuous latent space for GAN training. In GAN for sequences of discrete elements [11] and RELGAN [12], Gumbel-Softmax approaches are suggested to approximate the discrete-valued distribution for continuous-valued distribution.

\vspace{-0.3em}

\section{Conditional hybrid GAN}
We propose an end-to-end deep generative model for generating sequence conditioned on the context. The proposed C-Hybrid-GAN model is trained by considering the alignment relationship between attributes of the sequences and their corresponding context. The idea contains i) a generator with three independent relational memory based conditional sub-networks, and ii) a discriminator based on single relational memory for long-distance dependency modeling and for providing informative updates to the generator. Gumbel-Softmax relaxation technique is exploited to train GAN for generating discrete-valued sequences. Particularly, a hybrid structure is used in the adversarial training stage, containing three independent branches for attributes in the generator and one branch for concatenating attributes in the discriminator. Relational memory is employed to model not only the dependency inside sequence of attribute during the training of the generator but also the consistency among three sequences of attributes during the training of the discriminator.


\subsection{Relational memory core}
Relational memory core (RMC) as a relational reasoning technique proposed by Santoro et al.[13] is composed of a fixed set of memory slots and employs multi-head dot product attention (MHDPA) also known as self-attention [14] between the memory slots to enable interaction between them and facilitate long term dependency modeling. Santoro et al. [13] show empirically that RMC is better-suited for tasks such as language modeling that benefits from relational reasoning across the sequential information as compared to the LSTM.

Formally, we suppose $M_t$ represents the memory of the RMC module and $x_t$ represents the input at time $t$.
Let $H$ represent the number of attention heads. For each head $h$, $M_t$ is used to construct queries $Q_t^{(h)} = M_tW_q^{(h)}$, and its combination with $x_t$ is used to construct keys $K_t^{(h)} = [M_t;x_t]W_k^{(h)}$ and values $V_t^{(h)} = [M_t;x_t]W_v^{(h)}$, where $[;]$ represents the row-wise concatenation operation, $W_k^{(h)}, W_v^{(h)}, W_q^{(h)}$ are weights.
An attention weight is computed from $Q_t^{(h)}$ and $K_t^{(h)}$, and $\tilde{M}_{t+1}$ is computed as the product of attention weight and the value, as follows.
\begin{equation}
    \tilde{M}_{t+1} = [\tilde{M}_{t+1}^{(1)}:\cdots:\tilde{M}_{t+1}^{(H)}],
    \ \tilde{M}_{t+1}^{(h)} = softmax(\frac{Q_t^{(h)}(K_t^{(h)})^T}{\sqrt{d_k}})    \ V_t^{(h)}
\end{equation}
where $d_k$ is the column dimension of the key $K_t^{(h)}$ and $[:]$ represents column-wise concatenation.
Then, the memory $M_{t+1}$ is updated and the output $o_{t}$ is computed from $\tilde{M}_{t+1}$ and $M_t$ by
\begin{equation}
    M_{t+1} = f_{\theta_1}(\tilde{M}_{t+1}, M_t),
    \ o_t = f_{\theta_2}(\tilde{M}_{t+1}, M_t)
\end{equation}
where $f_{\theta_1}$ and $f_{\theta_2}$ are parameterized functions consisting of skip connections, multi-layer perceptron, and gated operation.

\subsection{Generator with three relational sub-networks}
The role of the generator network is to generate a sequence with given context information. The generator network is composed of three independent relational memory based conditional sub-networks. Each sub-network is responsible for generating a sequence of a particular attribute conditioned on the context, for example, in melody sequence generation from lyrics, i.e. either a pitch sequence, $\hat{y}^p_{1,\ \cdots,\ T}$, a duration sequence,  $\hat{y}^d_{1,\ \cdots,\ T}$, or a rest sequence, $\hat{y}^r_{1,\ \cdots,\ T}$. The key component of each sub-network is the RMC module.

\begin{figure*}[!ht]
   \centering
   \includegraphics[width=9.0cm]{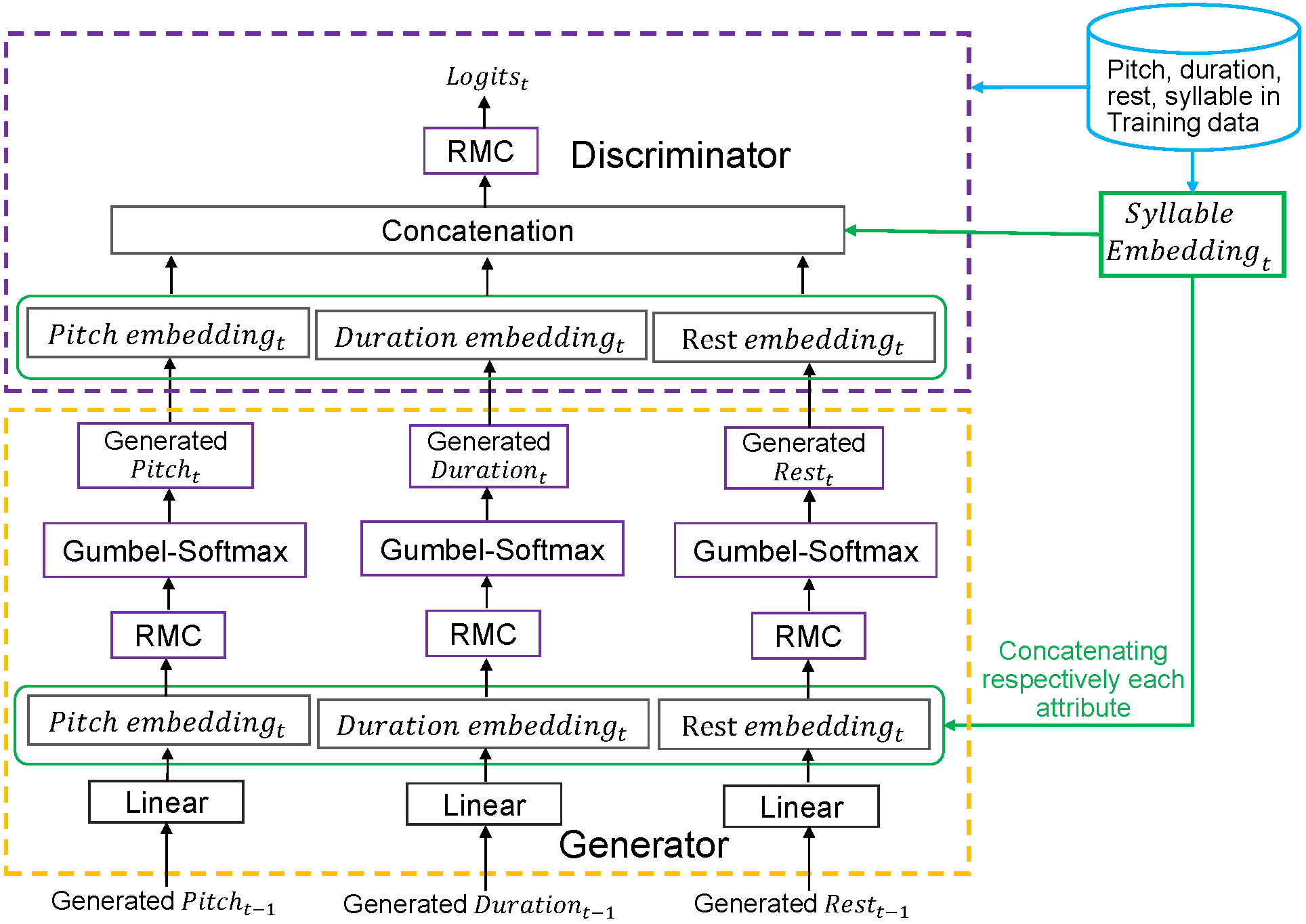}
    \caption{Architecture of conditional hybrid GAN.}
    \label{fig:arch}
\end{figure*}

 We take melody sequence generation from lyrics in Fig. \ref{fig:arch} to explain our generation process, which can be easily extended to other research scenario of sequence generation with multiple attributes.  Here, with the pitch attribute as an example, the similar process applies to the other two music attributes (duration, rest). At each time step $t$, the input to the sub-network is one-hot encoded representation of the pitch attribute generated during the previous time step $\hat{y}_{t-1}^p \in \mathbb{R}^{100}$ and the embedded lyrics syllable $x_t \in \mathbb{R}^{20}$. During the forward pass of the sub-network, $\hat{y}_{t-1}^p$ is passed through a linear layer to obtain a dense representation of the pitch attribute. The dense representation of the pitch attribute is then concatenated with $x_t$ and passed through a fully connected (FC) layer with ReLU activation. The output of FC layer and the RMC memory $M_{t-1}$ are then passed through the RMC layer. The RMC output is then passed through a linear layer to obtain the output logits $o_t \in \mathbb{R}^{100}$. The Gumbel-Softmax operation is performed on $o_t$ to obtain the one-hot approximation of the pitch attribute  $\hat{y}_t^p \in \mathbb{R}^{100}$. $\hat{y}_{t-1}^p \sim Uniform(0, 1)$ is used for  the initial time step.

Since sequences with length $T=20$ are utilized in our model, we repeat this process for $20$ steps and generate the pitch sequence $\hat{y}^p = [\hat{y}_1^p, \hat{y}_2^p, \cdots , \hat{y}_T^p]$ where $\hat{y}_t^p \in \mathbb{R}^{100}, 1 \le t \le 20$. The other two sub-networks respectively follow the same procedure to generate a duration sequence $\hat{y}^d = [\hat{y}_1^d, \hat{y}_2^d, \cdots , \hat{y}_T^d]$ where $\hat{y}_t^d \in \mathbb{R}^{12}, 1 \le t \le 20$ and a rest sequence $\hat{y}^r = [\hat{y}_1^r, \hat{y}_2^r, \cdots , \hat{y}_T^r]$ where $\hat{y}_t^r \in \mathbb{R}^{7}, 1 \le t \le 20$.

In the generator network, the embedding dimensions of pitch, duration, and rest are set to 32, 16, and 8 respectively. In the pitch sub-network, the fully connected layer following the embedding layer uses ReLU activation with 64 units. The RMC layer following the fully connected layer uses a single memory slot with the head size set to 64, the number of heads set to 2, and the number of blocks set to 2. In the duration sub-network, the fully connected layer following the embedding layer uses ReLU activation with 32 units. The RMC layer following the fully connected layer uses a single memory slot with the head size set to 32, the number of heads set to 2, and the number of blocks set to 2. In the rest sub-network, the fully connected layer following the embedding layer uses ReLU activation with 16 units. The RMC layer following the fully connected layer uses a single memory slot with the head size set to 16, the number of heads set to 2 and the number of blocks set to 2.

\subsection{Gumbel-Softmax}
Training GANs for the generation of discrete data faces a non-differentiable problem due to discrete-valued output from the generator. The gradient of the generator loss $\frac{\partial loss_G}{\partial \theta_G}$ cannot be back propagated to the generator via the discriminator and hence generator parameters $\theta_G$ cannot be updated. To overcome this issue, we apply the Gumbel-Softmax relaxation technique. Using the generator sub-network responsible for generating the pitch attribute as an example, we explain more about the non-differentiability issue. We know in our data the number of distinct MIDI numbers is 100, at time step $t$, denote the output logits obtained from generator sub-network as $o_t \in \mathbb{R}^{100}$, then we can obtain the next one-hot encoded pitch attribute $y_{t+1}^p$ by sampling:

\begin{equation}
    y_{t+1}^p \sim softmax(o_t).
\end{equation}

Here, $softmax(o_t)$ represents the multinomial distribution on the set of all possible MIDI numbers. Because the sampling operation in (3) is not differentiable, this implies the presence of a step function at the output of the sub-network. Since the derivative of a step function is 0, $\frac{\partial loss_G}{\partial \theta^p_G} = 0$, this is the non-differentiability issue mitigated by applying the Gumbel-Softmax relaxation. The Gumbel-Softmax relaxation defines a continuous distribution over the simplex that can approximate samples from a categorical distribution [15][16]. Applying the Gumbel-Softmax , we can reparameterize the sampling in (3) as

\begin{equation}
    \hat{y}_{t+1}^p = \sigma(\beta(o_t + g_t))
\end{equation}

where $\beta>0$ is a tunable parameter called \textit{inverse temperature}, \ $g_t^{(i)}$ is from the \textit{i.i.d} standard Gumbel distribution i.e. $g_t^{(i)} = -\log(-\log{U_t^{(i)}}) \ \text{with} \  U_t^{(i)} \sim Uniform(0, 1)$. As $\hat{y}_{t+1}^p$ in (4) is differentiable w.r.t. $o_t$, we can use it instead of $y_{t+1}^p$ as the input to the discriminator.
\subsection{Discriminator with single relational network}
The discriminator has a relational memory based network. Its role is to distinguish between the generated sequence and the real sequence conditioned on the context. We continue to take melody sequence generation from lyrics to explain the stage of the discriminator with single relational network. At each time step $t$, the input to the discriminator network is the one-hot encoded representation of each music attribute (either real or generated) i.e. the pitch attribute $y_t^p \in \mathbb{R}^{100}$, duration attribute $y_t^d \in \mathbb{R}^{12}$, and rest attribute $y_t^r \in \mathbb{R}^{7}$ and the embedded representation of lyrics syllable $x_t \in \mathbb{R}^{20}$.

Initially, during the discriminator network forward pass, each music attribute $y_t^p, y_t^d, y_t^r$ is independently passed through a linear layer to obtain a dense representation for each music attribute. The dense representation for each music attribute is concatenated together with $x_t$ to form a syllable conditioned triplet of music attributes \{$y_t^p, y_t^d, y_t^r$\}. We then pass the syllable conditioned triplet of music attributes \{$y_t^p, y_t^d, y_t^r$\} through a dense layer with ReLU activation. The outputs of the dense layer and the RMC memory $M_{t-1}$ are passed through the RMC layer. The RMC output is passed through a linear layer with a single unit to obtain the output logits $o_t \in \mathbb{R}$.

Since the length of sequences is $T=20$, we repeat this process for $20$ steps and generate a sequence of output logits $o = [o_1, o_2, \cdots ,o_T]$. We then take the mean of $o$ and use it for the loss computation. Let $o$ and $\hat{o}$ represent the output logits obtained for real and generated music attributes conditioned on lyrics are passed through the discriminator respectively. Then, the discriminator loss is given by


\begin{equation}
    loss_D = \log sigmoid(\frac{1}{T}\sum_{t=1}^T{o_t} - \frac{1}{T}\sum_{t=1}^T{\hat{o}_t})
\end{equation}

Here, we employ the relativistic standard GAN (RSGAN)[17] loss function. Intuitively, the loss function in (5) directly estimates the average probability that real melody is more realistic than generated melody. We simply can set the generator loss as $loss_G = -loss_D$.

In the discriminator network, the embedding dimensions of pitch, duration, and rest are set to 32, 16, and 8 respectively. The fully connected layer following the embedding layer uses ReLU activation with 64 units. The RMC layer following the fully connected layer contains a single memory slot with the head size, the number of heads, and the number of blocks set to 64, 2, and 2 respectively.



\vspace{-0.3em}
\section{Evaluation}
In this section, we discuss the experimental setup and experimental results to demonstrate the feasibility of our proposed C-Hybrid-GAN. To evaluate our proposed architecture, we use Self-BLEU [18] to measure generated sample diversity and maximum mean discrepancy (MMD)[19] to measure generated sample quality. The effect of lyrics conditioning is also investigated. Melody-lyrics aligned dataset used in [1] is utilized in our experiment, which contains 13,251 sequences, with each consisting of 20 syllables aligned with the triplet of music attributes \{$y_t^p, y_t^d, y_t^r$\}. The dataset is split into training, validation and testing sets with the ratio of 8:1:1. Conditional hybrid MLE (C-Hybrid-MLE) and conditional LSTM GAN (C-LSTM-GAN)[1] are used to compare with our proposed C-Hybrid-GAN.
\subsection{Experimental setup}
We use the Adam optimizer with $\beta_1=0.9$ and $\beta_2=0.99$. We perform gradient clipping if the norm of the gradients exceeds 5. Initially, the generator network is pre-trained with the MLE objective for 40 epochs with a learning rate of 1e-2. We then perform adversarial training for 120 epochs with a learning rate of 1e-2 for both the generator and discriminator. Each step of adversarial training is composed of a single discriminator step and a single generator step. The batch size is set to 512 and a maximum temperature $\beta_\text{max}=1000$ is used during the adversarial training.

\subsection{Diversity evaluation of generated sequences}

\begin{figure*}[!ht]
    \centering
    \includegraphics[width=10.0cm]{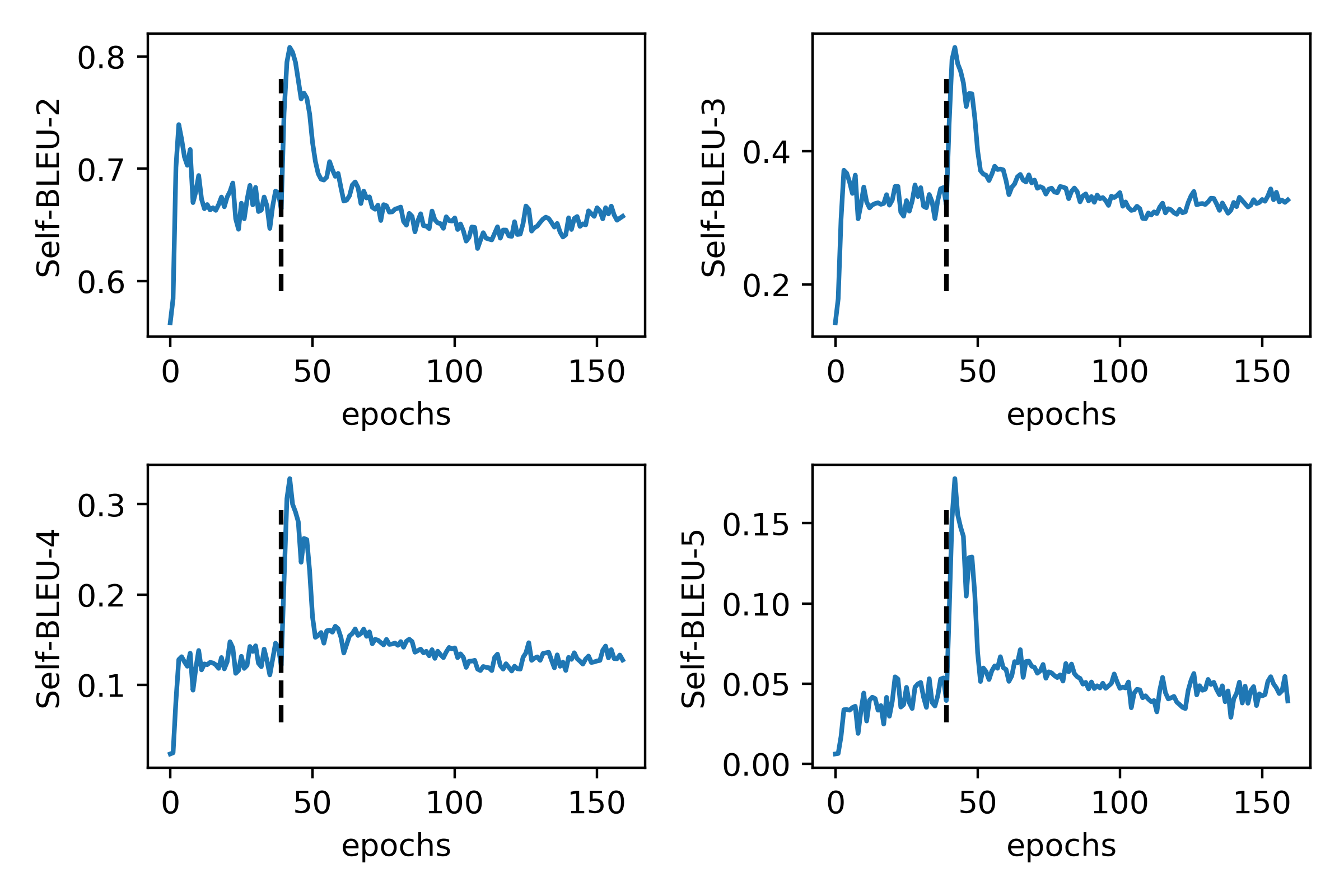}
    \caption{Training curves of self-BLEU scores on testing dataset.}
    \label{fig:self_bleu_trend}
\end{figure*}

We use the Self-BLEU [18] score as a means to measure the diversity of melodies generated by our proposed model. The value of the Self-BLEU score ranges between 0 and 1 with a smaller value of Self-BLEU implying a higher sample diversity hence a less chance of mode collapse in the GAN model.

Intuitively, the Self-BLEU score measures how a generated melody sample is similar to the rest of the generated melody samples. With respect to our proposed model, to compute the Self-BLEU score we first combine the pitch, duration, and rest sequences generated by each generator sub-network to form a sequence of music attributes i.e. a melody. As an example let us assume the sequences of pitches, durations, and rests generated by each corresponding sub-network is $\hat{p} = [\hat{p}_1, \hat{p}_2, \cdots ,\hat{p}_T], \ \hat{d} = [\hat{d}_1, \hat{d}_2, \cdots ,\hat{d}_T], \ \hat{r} = [\hat{r}_1, \hat{r}_2, \cdots ,\hat{r}_T]$ respectively. Then we can represent a melody as $\hat{n} = [\hat{p}_1\hat{d}_1\hat{r}_1, \hat{p}_2\hat{d}_2\hat{r}_2, \cdots ,\hat{p}_T\hat{d}_T\hat{r}_T]$.

To compute the Self-BLEU score, we regard one generated melody as the hypothesis and the rest of the generated melodies as the references. We calculate the BLEU score for every generated melody and define the average BLEU score to be the value of the Self-BLEU metric. The results of Self-BLEU are shown in Fig. \ref{fig:self_bleu_trend}. During the adversarial training, Self-BLEU values of our C-Hybrid-GAN architecture reach the peak around 45 epochs, degrade until 100 epochs, and then approach to the stability. The results indicate that the diversity of generated melody samples gets better with the decrease of Self-BLEU and keep unchanged from 100 epochs to 150 epochs.


\subsection{Quality evaluation of generated sequences}

\begin{figure*}[!ht]
    \centering
    \includegraphics[width=10.0cm]{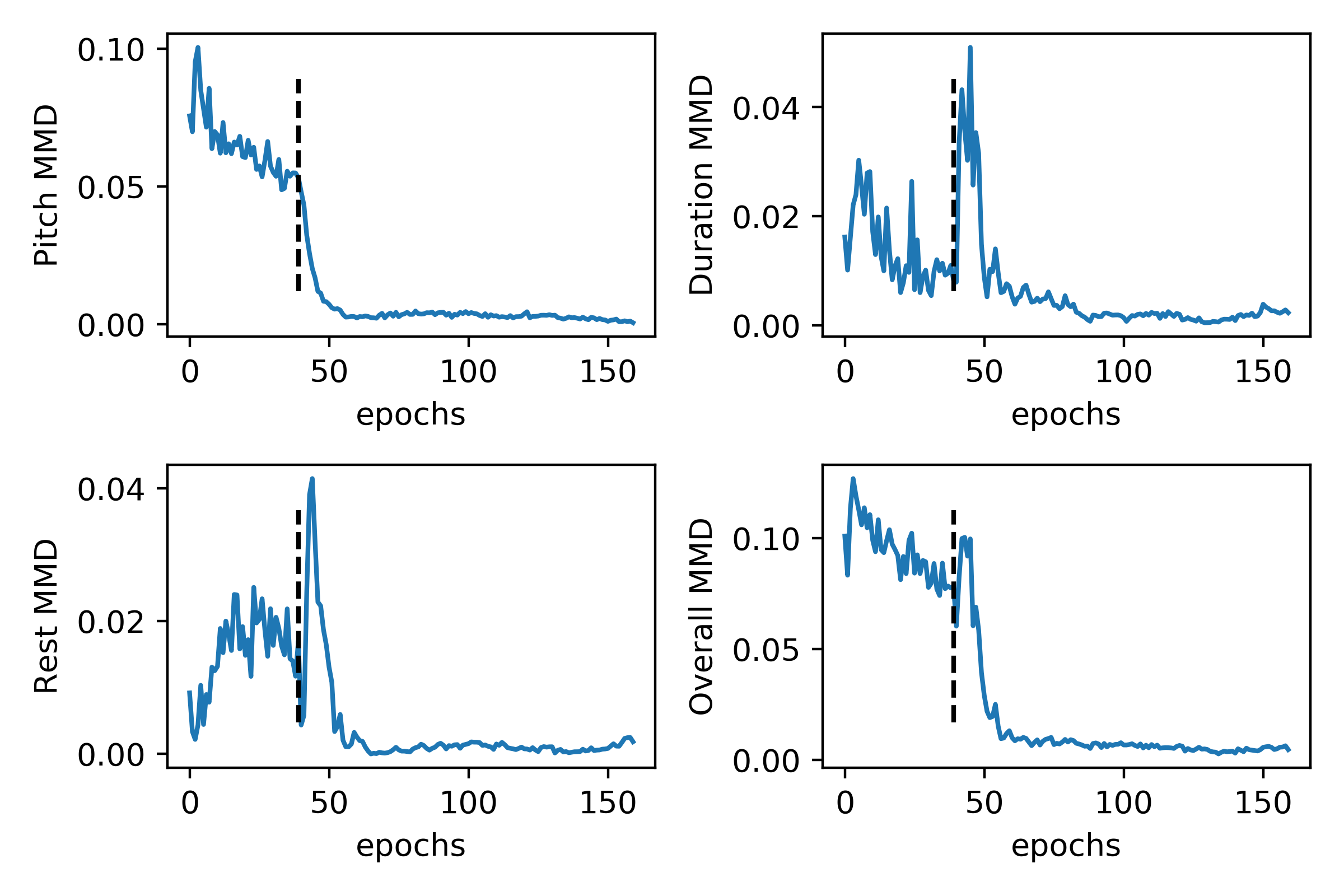}
    \caption{Training curves of of MMD scores on testing dataset.}
    \label{fig:mmd_trend}
\end{figure*}

The quality of generated melodies is investigated using a MMD [19] unbiased estimator. The smaller MMD value indicates a better performance. As shown in Fig. \ref{fig:mmd_trend}, at each epoch, the generator outputs a sequence of pitches, a sequence of durations, and a sequence of rests. Using these generated sequences of pitchs, durations, and rests together with the corresponding real sequences of pitches, durations, and rests we can compute MMD values of pitch, duration and rest respectively. The sum of these three values can be utilized to obtain the overall MMD of the testing set.




During the adversarial training, we can see the sample quality, as measured by the MMD, for the proposed C-Hybrid-GAN model, first increases with the quick decrease of MMD value until 50 epochs and then starts to approach the stability and keep unchanged until 150 epochs. Each trend indicated from MMD values of pitch, duration, or rest is consistent with that of the other two and the overall trends of MMD value. The results demonstrate that the overall quality of generated melodies is high as it has a low value of MMD.





\subsection{Effect of context conditioning}

\begin{figure*}
\subfloat[Note duration attribute: $d = 0.987$ is highlighted in red in each boxplot. Mean values are $\mu_{\text{rs}} = 1.008$, $\mu_{\text{rn}} = 1.007$ and $\mu_{\text{rns}} = 1.008$ respectively.]{\protect\includegraphics[height=3.4cm]{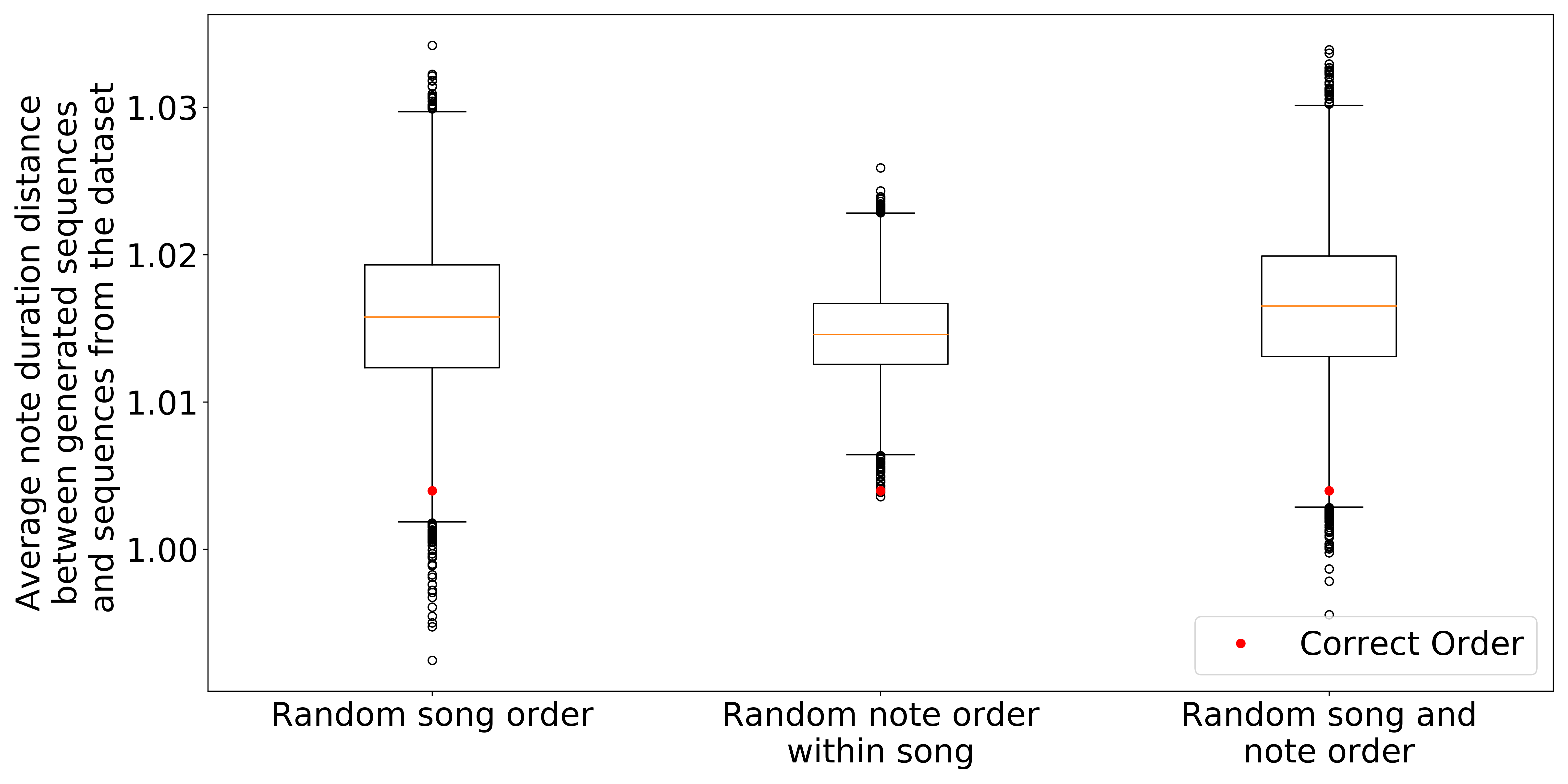}}
\hfill{}
\subfloat[Rest duration attribute: $d = 1.261$ is highlighted in red in each boxplot. Mean values are $\mu_{\text{rs}} = 1.367$, $\mu_{\text{rn}} = 1.370$ and $\mu_{\text{rns}} = 1.368$ respectively.]{\protect\includegraphics[height=3.4cm]{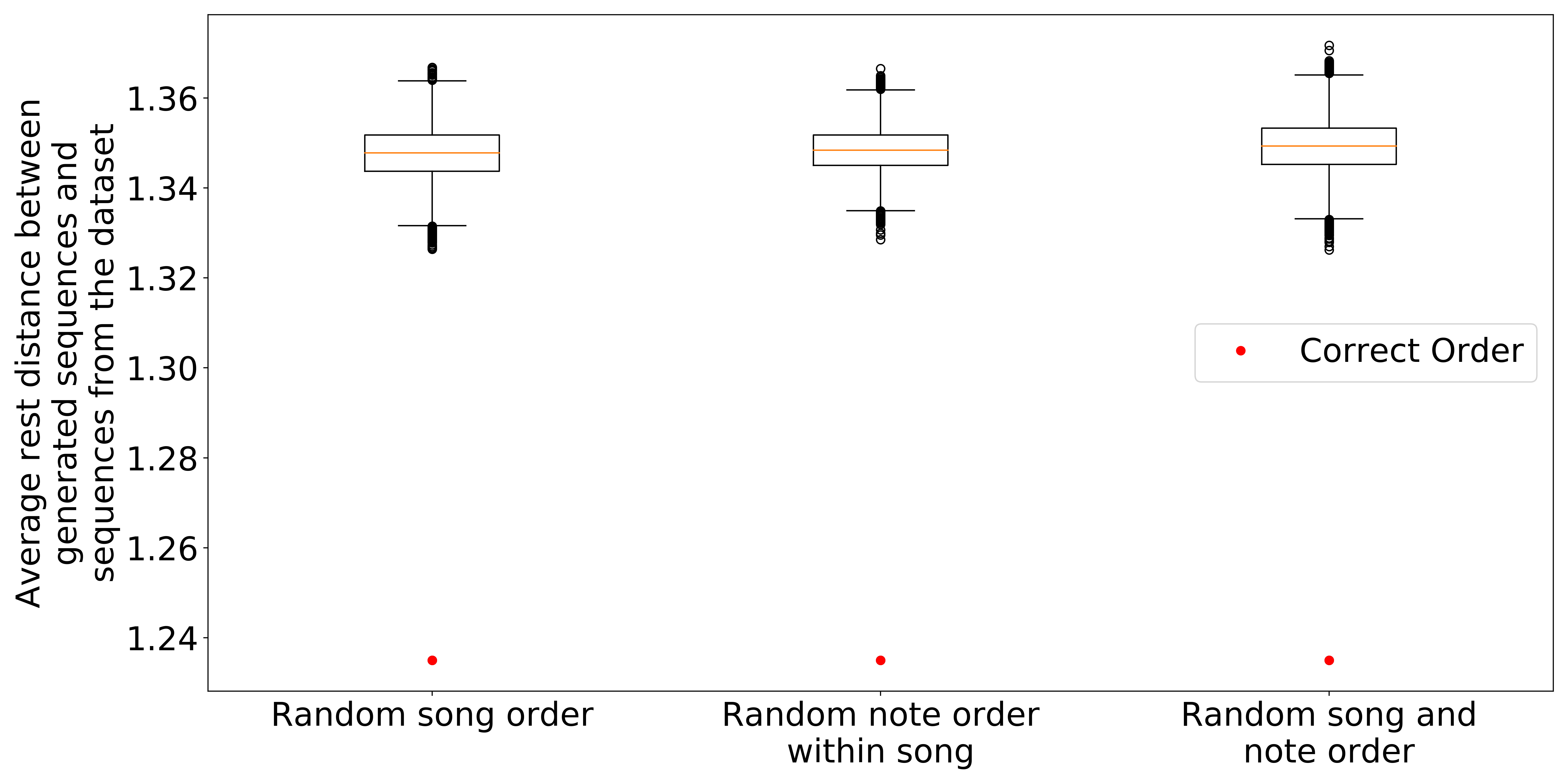}}
\protect\caption{\label{fig:duration_distance_boxplot}Boxplots of the distributions of $d_{\text{rs}}$, $d_{\text{rn}}$ and $d_{\text{rns}}$.}
\end{figure*}

To show the generated melodies are efficiently conditioned by lyrics, we follow the previous evaluation method proposed in [1], where an effect on the generated note duration and rest duration is studied. Average note duration distance between generated sequences and sequences from ground truth dataset is calculated in Fig. \ref{fig:duration_distance_boxplot}(a). Average rest duration distance between generated sequences and sequences from ground truth dataset is calculated in Fig. \ref{fig:duration_distance_boxplot}(b). The subscripts $_{\text{rs}}$, $_{\text{rn}}$ and $_{\text{rns}}$ respectively denote ``random songs'', ``random notes'', and ``random notes + songs''. In this experiment, $d$ is a real value, which is compared to the distribution of the random variables $d_{\text{rs}}$, $d_{\text{rn}}$ and $d_{\text{rns}}$, with $N = 1,051$ (number of songs in testing set) and $T=20$.


The three distributions are estimated using 10,000 samples for each random variable. As the results shown in Fig. \ref{fig:duration_distance_boxplot}, in each case, $d$ is statistically lower than the mean value, indicating that the generator learns useful correlation between syllable embeddings and note/rest durations. For a detailed evaluation method of lyrics conditioning refers to [1].

\subsection{Comparing with state-of-the-art methods}
To study if C-Hybrid-GAN can generate sequences that resemble the same distribution as training samples, quantitative evaluation is performed to compare existing state-of-the-art approaches following the previous quantitative measurements in [1], for example, 2-MIDI numbers repetitions, 3-MIDI numbers repetitions, MIDI numbers span, the number of unique MIDI, the number of notes without rest, average rest value within song, and song length. More detailed descriptions for these measurements can refer to [1].
\begin{table}
  \caption{Metrics evaluation of attributes.}
  \label{song_attr_table}
  \centering
  \resizebox{\textwidth}{!}{%
      \begin{tabular}{lcccc}
        \toprule
        & Ground Truth
        & \shortstack{C-LSTM-GAN}
        & \shortstack{C-Hybrid-GAN}
        & \shortstack{C-Hybrid-MLE}\\
        \midrule
        2-MIDI numbers repetitions & 7.4 & 7.7 & 6.6 & 6.3  \\

        3-MIDI numbers repetitions &  3.8 & 2.9 & 2.8 & 2.4  \\

        MIDI numbers span & 10.8 & 7.7 & 12.0 & 13.5  \\

        Number of unique MIDI numbers & 5.9 & 5.1 & 6.0 & 6.2 \\

        Average rest value within song & 0.8 & 0.6 & 0.7 & 1.1  \\

        Number of notes without rest & 15.6 & 16.7 & 15.8 & 13.2  \\

        Song length & 43.3 & 39.2 & 43.2 & 51.9  \\
        \bottomrule
      \end{tabular}%
  }
\end{table}

\begin{figure*}[!ht]
    \centering
    \includegraphics[width = \textwidth]{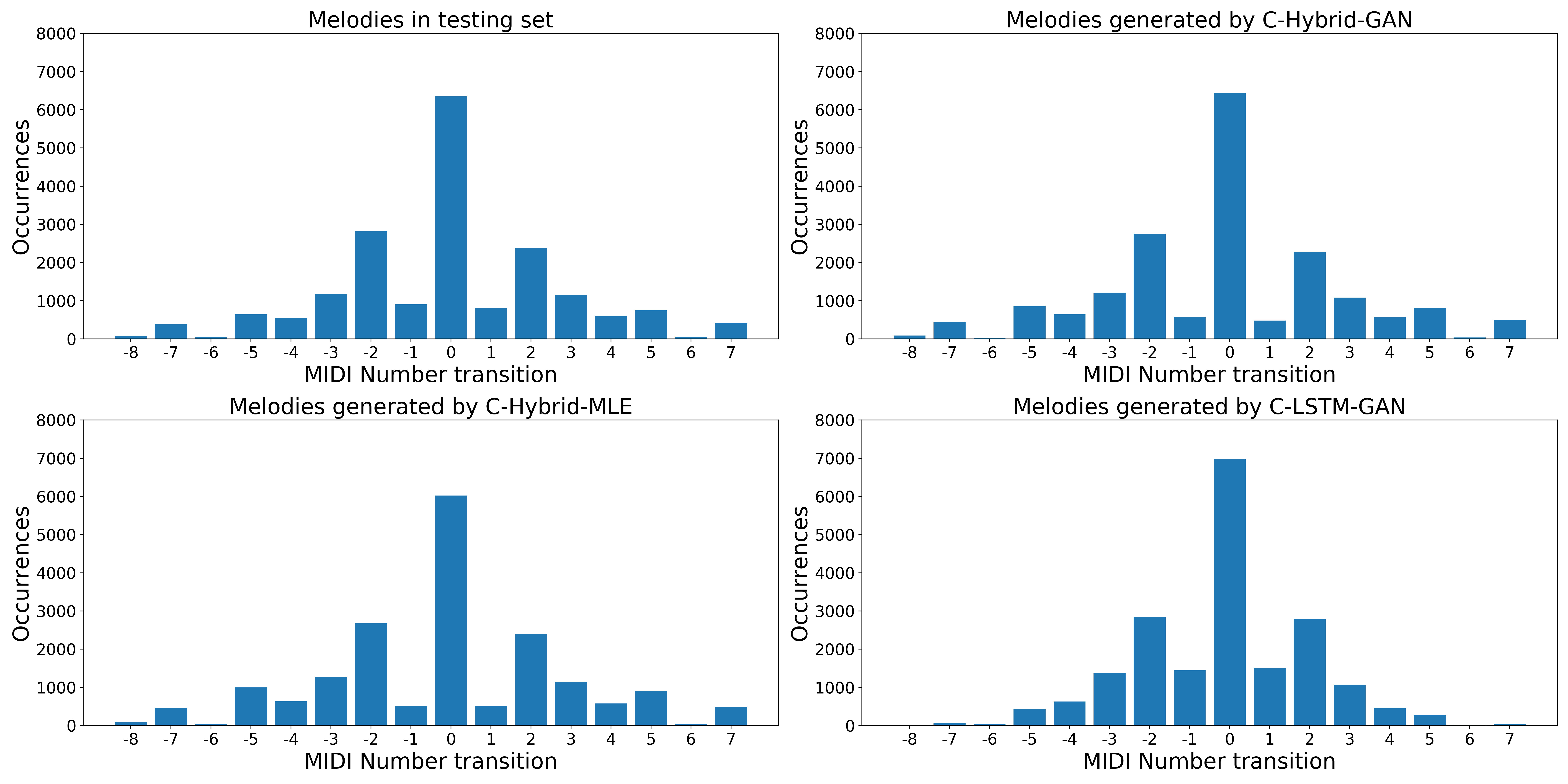}
    \caption{Distribution of transitions.}
    \label{fig:transitions}
\end{figure*}

Table \ref{song_attr_table} shows the results related to quantitative evaluation of generated melodies. it is very obvious that the proposed C-Hybrid-GAN architecture outperforms other competitive methods in most aspects. For pitch-related attributes such as MIDI number span and number of unique MIDI numbers, the proposed C-Hybrid-GAN model is closest to the corresponding value of the ground truth. In addition, for metrics on temporal attributes such as average rest value and the number of notes without rest, C-Hybrid-GAN is closest to the corresponding value of the ground truth.

Besides metrics discussed in Table \ref{song_attr_table}, the distribution of the transitions between MIDI numbers is very important attribute for quantitatively measuring generated melodies. Fig. \ref{fig:transitions} shows the distributions of the transitions for the melodies generated by our model C-Hybrid-GAN, C-Hybrid-MLE, and C-LSTM-GAN. According to the occurrence of MIDI number transition in the figures, it is very clear that the proposed C-Hybrid-GAN model can represent and capture well the distribution of MIDI number transition. This concludes our proposed model outperforms other competitive methods and best approximates the MIDI number transition in the ground truth.

%


\vspace{-0.5em}
\section{Conclusion}
Sequence generation from context has been an interesting research topic in the area of artiﬁcial intelligence. The goal of this generation is to design generative models that can automatically infer sequence when given context in a way similar to the human way. However, the current state-of-the-art generative models have the incapability of generating discrete-valued sequences with multiple attributes when given context.

In this paper, we propose the novel conditional hybrid generative adversarial network for generating sequence from context. Three independent discrete-valued sequences containing different attributes are exploited to learn context-conditioned sequence generation. In particular, the relational reasoning method is employed to learn the dependency inside independent sequence of the specific attribute during the training stage of the generator as well as the consistency among all independent sequences of attributes during the training stage of the discriminator. To avoid the non-differentiability problem in GANs for discrete data generation, we exploit the Gumbel-Softmax to approximate the distribution of discrete-valued sequences. Through extensive experiments of melody generation from lyrics including diversity and quality of generated sequence samples, effect of lyrics-based context conditioning, and comparison with existing works, we indicate that the proposed C-Hybrid GAN outperforms the existing cutting-edge methods in context-conditioned sequence generation with multiple attributes.

\section*{References}

[1] Yu, Y., Srivastava, A. \ \& Canales, S.\ (2019) Conditional LSTM-GAN for Melody Generation from Lyrics. arXiv:1908.05551 [cs.AI].

[2] Deng, K., Fei, T., Huang, X., \ \& Peng, Y.\ (2019) {IRC-GAN:} Introspective Recurrent Convolutional {GAN} for Text-to-video Generation, Proceedings of the Twenty-Eighth International Joint Conference on Artificial Intelligence, {IJCAI} 2019, pages 2216-2222.

[3] Tuan, Y.-L., \ \& Lee, H.-Y. \ (2018) Improving Conditional Sequence Generative Adversarial Networks by Stepwise Evaluation, arXiv:1808.05599 [cs.CL].

[4] Fedus, W., Goodfellow, I., Dai, A. M. \ (2018) MaskGAN: Better Text Generation via Filling in the\----\, arXiv:1801.07736 [stat.ML].

[5] Lin, K., Li, D., He, X., Zhang, Z., \ \& Sun, M.-T. \ (2017) Adversarial Ranking for Language Generation, arXiv:1705.11001 [cs.CL].

[6] Guo, J., Lu, S., Cai, H., Zhang, W., Yu, Y., \ \& Wang, J. \ (2017) Long Text Generation via Adversarial Training with Leaked Information, arXiv:1709.08624 [cs.CL].

[7] Yu, L., Zhang, W., \ \& Yu, Y. \ (2016) Seqgan:Sequence Generative Adversarial Nets with Policy Gradient, arXiv:1609.05473 [cs.LG].

[8]Chen, L., Dai, S., Tao, C., Zhang, H., Gan, Z., Shen, D., Wang, G, Zhang, R., \ \& Carin, L. \ (2018) Adversarial Text Generation via Feature-Mover's Distance, Advances in Neural Information Processing Systems 31: Annual Conference on Neural Information Processing Systems 2018  (NeurIPS), pages 4671-4682.

[9] Zhang, Y., Gan, Z., Fan, K., Chen, Z., Henao, R., \ \& Shen, D. \ (2017) Adversarial Feature Matching for Text Generation, arXiv:1706.03850 [stat.ML].

[10] Zhao, J., Kim, Y., Zhang, K., Rush, A. M., \ \& LeCun, Y. \ (2017) Adversarially Regularized Autoencoders, arXiv:1706.04223 [cs.LG].

[11] Kusner, M. J., \ \& Hernández-Lobato, J. M., \ (2016) GANS for Sequences of Discrete Elements with the Gumbel-softmax Distribution, arXiv:1611.04051 [stat.ML].

[12] Nie, W., Narodytska, N., \ \& Patel, A., \ (2019) RelGAN: Relational Generative Adversarial Networks for Text Generation, 7th International Conference on Learning Representations, {ICLR} 2019.

[13] Santoro, A., Faulkner, R., Raposo, D., Rae, J., Chrzanowski, M.,
Weber, T., Wierstra,D., Wierstra, O., Pascanu, R., \ \& Lillicrap, T., \ (2018) Relational recurrent neural networks, In Neural Information Processing Systems 2018 (NeurIPS), pages 7310-7310.

[14] Vaswani, A., Shazeer, N., Parmar, N., Uszkoreit, J., Jones, L.,
Gomez, A. N., Kaiser, Ł., \ \& Polosukhin, I., \ (2017) Attention is all you
need. In Advances in Neural Information Processing Systems 2017 (NeurIPS), pages 5998-6008.

[15] Jang, E., Gu, S., \ \& Poole, B., \ (2016) Categorical reparameterization with
gumbel-softmax, arXiv:1611.01144, 2016.

[16] Maddison, C. J., Mnih, A., \ \& Teh, Y. W., \ (2016) The concrete distribution:
A continuous relaxation of discrete random variables, arXiv:1611.00712, 2016.

[17] Jolicoeur-Martineau, A., \ (2018) The relativistic discriminator: a key element
missing from standard gan, arXiv:1807.00734, 2018.

[18] Zhu, Y., Lu, S., Zheng, L., Guo, J., Zhang, W., Wang, J., \ \& Yu, Y., \ (2018) Texygen: A Benchmarking Platform for Text Generation Model. In Special Interest Group on Information Retrieval (SIGIR) 2018, pages 1097-1100.

[19] Smola, A., Gretton, A., Song, L., \ \& Schölkopf, B., \ (2007) A hilbert space embedding for distributions, in Algorithmic Learning Theory (ALT), Springer Berlin Heidelberg,2007, pages 13-31.

\end{document}